\documentclass{article}
\usepackage[preprint]{spconf}
\copyrightnotice{\copyright\ IEEE 2021}
\toappear{To appear in {\it Proc.\ ICASSP2021, June 06-11, 2021, Toronto, Ontario, Canada}}
\usepackage{amsmath,graphicx}
\usepackage{booktabs}
\usepackage{url}
\usepackage[numbers,sort&compress]{natbib}
\ninept
\title{Transformer Language Models with LSTM-based Cross-utterance Information Representation}
\name{G. Sun, C. Zhang, P. C. Woodland \thanks{G. Sun is funded by the Cambridge Trust}}
\address{Cambridge University Engineering Dept., Trumpington St., Cambridge, CB2 1PZ U.K.\\
\small{\texttt{\{gs534,cz277,pcw\}@eng.cam.ac.uk}}}
\begin{document}
%

\maketitle
\begin{abstract}
The effective incorporation of cross-utterance information has the potential to improve language models (LMs) for automatic speech recognition (ASR).
To extract more powerful and robust cross-utterance representations for the Transformer LM (TLM), this paper proposes the R-TLM which uses hidden states in a long short-term memory (LSTM) LM. To encode the cross-utterance information, the R-TLM incorporates an LSTM module together with a segment-wise recurrence in some of the Transformer blocks. In addition to the LSTM module output, a shortcut connection using a fusion layer which bypasses the LSTM module is also investigated. The proposed system was evaluated on the AMI meeting corpus, the Eval2000 and the RT03 telephone conversation evaluation sets. The best R-TLM achieved 0.9\%, 0.6\% and 0.8\% absolute WER reductions over the single-utterance TLM baseline, and 0.5\%, 0.3\%, 0.2\% absolute WER reductions over a strong cross-utterance TLM baseline on the AMI evaluation set, Eval2000 and RT03 respectively. Improvements on Eval2000 and RT03 were further supported by significance tests. R-TLMs were found to have better LM scores on words where recognition errors are more likely to occur. The R-TLM WER can be further reduced by interpolation with an LSTM-LM.
\end{abstract}

\vspace{-0.2cm}
\section{Introduction}
\label{sec:intro}
A language model (LM) estimates the probability of a word sequence which is often decomposed into a product of conditional word prediction probabilities using the chain rule. LMs are widely used in many machine learning tasks, such as natural language understanding, machine translation and automatic speech recognition (ASR). In ASR, high performance LMs are found to be crucial to achieving good performance for both traditional noisy source-channel model systems \cite{Cmap, LMinASR2, AndrewLiuRNNLM} and more recent end-to-end systems \cite{LMinASR3, LMinASR4, LMinASR5}. While traditional $n$-gram LMs only provide context information from a small fixed number of 
preceding words \cite{Cngram}, neural network-based LMs including the recurrent neural network (RNN) LM (RNNLM), the long short-term memory (LSTM) LM \cite{LSTM, LSTMLM1, LSTMLM2}, and the Transformer LM (TLM) \cite{Transformer,TransformerXL,AACHEN1}, can provide richer context information from the full history of an utterance and achieve better ASR performance. More recently, context information from both past and future utterances has been taken into account in language modelling \cite{TransformerXL,MShumanparity,CrossuttPaper}.

Over the years, improvements have been found by exploring more powerful and robust cross-utterance representations for RNNLMs and TLMs separately. 
Cross-utterance RNNLMs were usually improved by including richer information into a compact vector. 
In \cite{C8}, an LM was trained without resetting the hidden states at utterance boundaries so that a longer history can be encoded. Alternatively, global and local topic vectors, and neural-based cache models were integrated into LMs \cite{C3,C6,C7}. 
More recently, an extra neural network component, such as a hierarchical RNN or a pre-trained LM \cite{C11}, was used to encode the cross-utterance information into a vector representation for LM adaptation \cite{C9,C10,Cdeepcont}. 
On the other hand, improvements in cross-utterance TLMs were mainly from efficient extension of attention spans, such as using segment-wise recurrence between two adjacent segments \cite{TransformerXL}, adopting adaptive attention spans, or applying specially-designed masks to cope with much longer input sequences \cite{ADAPTIVESPAN, STRIDEDATTEN}. It has been found that such TLMs can reduce ASR word error rates (WERs) via LM rescoring \cite{AACHEN1, AACHEN2,MSLong}. 
Moreover, an RNN structure can also be used in a TLM to enhance its local correlations  \cite{LOCALRNN,LOCALRNN2}.

While the performance of both RNNLMs and TLMs can be improved by taking into account cross-utterance information, methods to incorporate such information are potentially complementary. 
To introduce the complementary LSTM hidden states into TLMs, the R-TLM structure (R stands for recurrence) is proposed which integrates an LSTM module into the TLM. The LSTM module contains a single layer and an optional shortcut connection, and LSTM hidden states are carried over from the preceding segment. The proposed LSTM module can be used in conjunction with the segment-wise recurrence from Transformer-XL \cite{TransformerXL} that leads to the R-TLM XL structure.
The proposed R-TLM and R-TLM XL structures are evaluated using the AMI meeting corpus, the Eval2000 and the RT03 telephone conversation evaluation sets, where consistent improvements in word error rate (WER) were found. Further reductions in WER were found by interpolating an R-TLM with an LSTM-LM.

The rest of this paper is organised as follows. Section~\ref{sec:tlm} reviews the TLM and the Transformer XL structures. Section~\ref{sec:xutttlm} describes the proposed R-TLM. The experimental setup and results are given in Sec.~\ref{sec:setup} and \ref{sec:results} respectively, followed by the conclusion in Sec.~\ref{sec:conc}.

\section{Transformer and Transformer-XL LMs}
\label{sec:tlm}

\begin{figure}[h]
    \centering
    \includegraphics[scale=0.45]{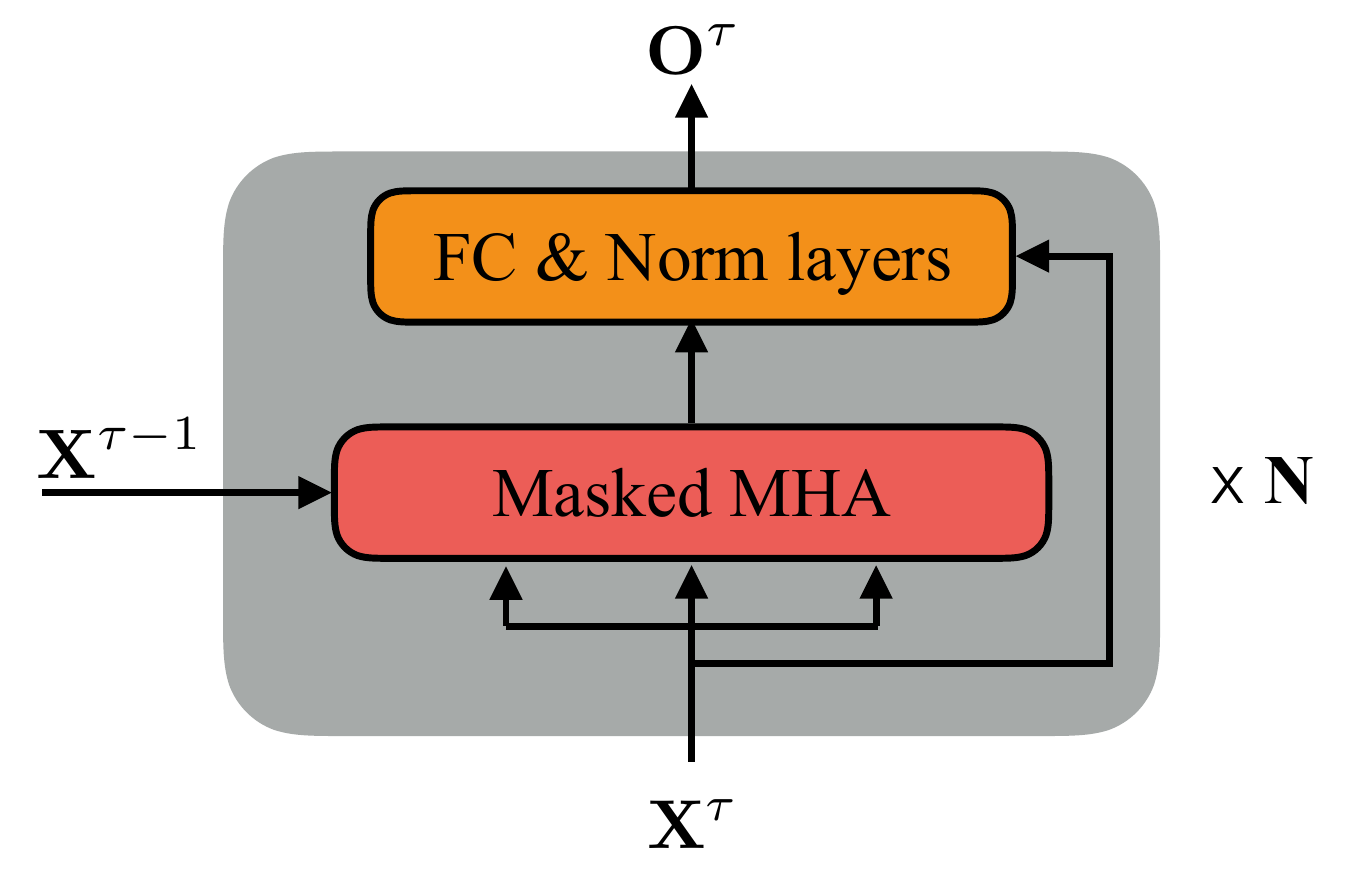}
    \vspace{-0.3cm}
    \caption{Transformer (w/o $\mathbf{X}^{\tau-1}$) and Transformer-XL (with $\mathbf{X}^{\tau-1}$) structures with $N$ repeated blocks. The MHA layer uses upper-triangular masks. ``Norm layers'' stands for layer normalisation.}
    \label{fig:tlm}
    \vspace{-0.5cm}
\end{figure}
The uni-directional TLM uses the modified decoder part of the full Transformer model in \cite{Transformer} as shown in Fig \ref{fig:tlm}. The model consists of a stack of blocks where each block includes a masked multi-head attention (MHA) module with a residual connection and a set of fully-connected (FC) layers. Taking segments with a fixed number of words as the LM input during training, the computation procedure of the masked MHA module in the $l$-th layer at segment $\tau$ is 
\begin{align}
\label{trans1}
    \mathbf{Q}^\tau_i&= \mathbf{X}^\tau\mathbf{W}_i^Q,\ \mathbf{K}^\tau_i =  \mathbf{X}^\tau\mathbf{W}_i^K,\ \mathbf{V}^\tau_i =  \mathbf{X}^\tau\mathbf{W}_i^V,\\
    \label{trans2}
    \textbf{H}^\tau_i&= \text{Softmax}({(\mathbf{Q}^\tau_i{\mathbf{K}_i^\tau}^{\text{T}}})/{\sqrt{d_k}})\mathbf{V}^\tau_i,\\
    \label{trans3}
    \mathbf{\tilde O}^\tau&= \mathbf{X}^\tau + \text{Concat}(\textbf{H}^\tau_1, \textbf{H}^\tau_2,...,\textbf{H}^\tau_n)\mathbf{W}^O,
\end{align}
where $\mathbf{X}^{\tau}=[\mathbf{x}_1^\tau, \mathbf{x}_2^\tau,...,\mathbf{x}_T^\tau]$ is the concatenation of the input vector (word embedding) sequence in the $\tau$-th utterance with $T$ input words. $\mathbf{W}_i^Q, \mathbf{W}_i^K, \mathbf{W}_i^V$ are model parameter matrices which transform the input sequence into the \textit{query}, \textit{key}, and \textit{value}. $d_k$ is the dimension of the key vectors. The Concat($\cdot$) function concatenates the output of each head to form a longer vector which is transformed into the output of the MHA, $\mathbf{\tilde O}^\tau$, using the parameter matrix $\mathbf{W}^O$ and a shortcut connection to the input $\mathbf{X}^\tau$. 
The final output $\mathbf{O}^\tau$ of the current block is denoted as $\text{TransformerBlock}(\mathbf{X}^{\tau})$ and generated by forwarding $\mathbf{\tilde O}^\tau$ through a stack of FC layers as
\begin{equation}
    \mathbf{O}^\tau=\text{FFN}(\mathbf{\tilde O}^\tau) + \mathbf{\tilde O}^\tau,
    \label{trans4}
\end{equation}
where FFN($\cdot$) denotes a feed-forward network. The output of the current block, which has the same dimension as $\mathbf{X}^\tau$, becomes the input to the next block until the  output layer. To indicate relative positions of tokens in a segment, relative positional encoding representing the index difference between the vector in the query sequence and the vector in the key sequence, can be included when calculating the dot-product in Eqn. (\ref{trans2}), as described in \cite{relativepos, TransformerXL}. 

To leverage more information by taking into account segments preceding the current one, segment-wise recurrence \cite{TransformerXL} is designed to append the input sequence to each TLM block of the current segment with the inputs from the previous segment. That is
\begin{equation*}
    \mathbf{\tilde X}^\tau = [\text{sg}(\mathbf{x}_1^{\tau-1}), \text{sg}(\mathbf{x}_2^{\tau-1}),...,\text{sg}(\mathbf{x}_T^{\tau-1}), \mathbf{x}_1^\tau, \mathbf{x}_2^\tau,...,\mathbf{x}_T^\tau],
\end{equation*}
where $\text{sg}(\cdot)$ denotes the stop-gradient operation which prevents gradients from back propagating through the appended history input. The remaining procedure is the same as the Transformer except that only the outputs corresponding to the current segment will be passed to the next block. The Transformer-XL is denoted as $\text{TransformerBlock}(\mathbf{X}^{\tau},\text{sg}(\mathbf{X}^{\tau-1}))$ or TLM XL, where XL stands for the use of segment-wise recurrence for extra-long sequences.

The TLM and TLM XL are trained with the cross-entropy loss. To match the training condition with segments of $T$ words at test-time, the start of each utterance in the LM rescoring stage is extended to form a $T$-word segment by concatenating the 1-best ASR output hypotheses from past utterances, denoted as the \emph{extended history}. When segment-wise recurrence is used for rescoring, output vectors from each Transformer block of the preceding $T$-word segment will also be used when computing the current segment. 

\section{R-TLM Structure}
\label{sec:xutttlm}
\begin{figure}[h]
    \centering
    \includegraphics[scale=0.45]{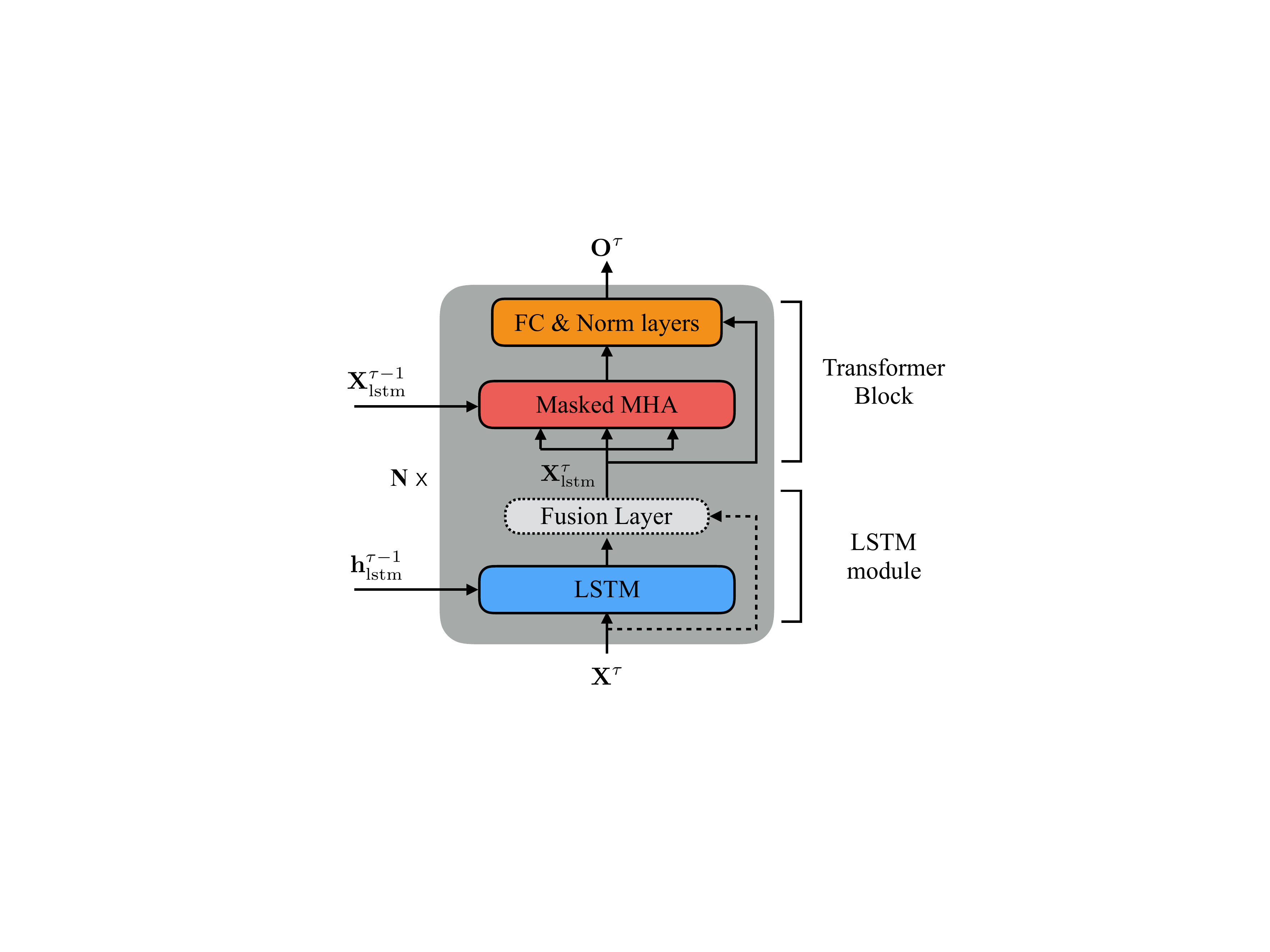}
    \vspace{-0.5cm}
    \caption{R-TLM structure where one of the $N$ repeated blocks is shown. The Transformer block is the same as in Fig. \ref{fig:tlm}, and the LSTM module is added to only a subset of Transformer blocks. The fusion layer and the shortcut in the LSTM module are optional. }
    \label{fig:lstmtlm}
    \vspace{-0.3cm}
\end{figure}

To make use of the complementary nature of the LSTM long-term representation and the segment-wise recurrence in the TLM XL, 
an LSTM module is added to a subset of Transformer blocks 
as shown in Fig. \ref{fig:lstmtlm}. 
For a Transformer block that includes the LSTM module, the input sequence $\mathbf{X}^\tau$ will be processed by a single-layer LSTM first before being sent to the masked MHA. The hidden states of the LSTM, including the output state and the memory cell state, are initialised by the final hidden states of the past segment $\tau-1$. The LSTM output sequence will be either sent to the MHA directly, or combined with the LSTM input through a fusion layer to form a shortcut connection that bypasses the LSTM layer. The computation procedure can be summarised as
\begin{align}
    \mathbf{X}^{\tau}_{\text{lstm}}, \mathbf{h}^{\tau}_{\text{lstm}}&= \text{LSTM}(\mathbf{X}^\tau, \text{sg}(\mathbf{h}^{\tau-1}_{\text{lstm}})),\\
    \label{eq:proj1}
    \mathbf{\tilde X}^{\tau}_{\text{lstm}}&= \text{Fusion}(\mathbf{X}^{\tau}_{\text{lstm}}, \mathbf{X}^{\tau}),\\
    \mathbf{O}^{\tau}&= \text{TransformerBlock}(\mathbf{\tilde X}^{\tau}_{\text{lstm}}),
\end{align}
where $\mathbf{X}^{\tau}_{\text{lstm}}$ and $\mathbf{\tilde X}^{\tau}_{\text{lstm}}$ are the outputs of the LSTM and the fusion layer respectively, 
and Fusion($\cdot$) refers to the combination operation in the optional fusion layer. The TransformerBlock($\cdot$) operation implements Eqns. \eqref{trans1}--\eqref{trans4}. When segment-wise recurrence is applied to the MHA module during training, the extended input will be drawn from the LSTM output of the last segment as 
\begin{equation}
    \mathbf{O}^{\tau} = \text{TransformerBlock}(\mathbf{\tilde X}^{\tau}_{\text{lstm}}, \text{sg}(\mathbf{\tilde X}^{\tau-1}_{\text{lstm}})),
    \label{eq:proj4}
\end{equation}
which guarantees that the vectors added in MHA have gone through the same computation procedure.
Meanwhile, in the fusion layer, each output vector of the LSTM layer is concatenated with the input vector at the corresponding position, and the fused vectors are passed through an FC layer, as in Eqn. \eqref{eq:proj5} shown below.
\begin{equation}
    \mathbf{\tilde X}^{\tau}_{\text{lstm}} = f(\mathbf{W}^c\mathbf{X}^{\tau}_{\text{lstm}} + \mathbf{U}^c\mathbf{X}^{\tau} + \mathbf{b}^c),
    \label{eq:proj5}
\end{equation}
where $\mathbf{W}^c$, $\mathbf{U}^c$ and $\mathbf{b}^c$ are model parameters, f($\cdot$) represents any activation functions among which the linear activation and the rectified linear unit (ReLU) were used in this paper. The training and rescoring procedure is the same as for TLM XL. 

The advantages of using the R-TLM structure are as follows. First, during the rescoring stage at test-time, there usually exists word errors in the transcriptions of past utterances, and the LSTM layer in the R-TLM structure provides LSTM hidden state representations which are more robust against such errors \cite{ErrorAnalysis}. Second, the R-TLM provides complementary history representations from both the LSTM and Transformer-XL to the attention-based representation, and increases the network capability. Third, since empirically LSTM-LMs usually perform better than TLMs on small datasets, the R-TLM, as a combination of the two, can potentially perform more consistently regardless of the size of the LM training set.

\section{Experimental Setup}
\label{sec:setup}
\vspace{-0.3cm}
\subsection{Data Sets}
The R-TLM was evaluated on the AMI meeting transcription \cite{AMI} and the Switchboard (SWB) telephone conversational transcription task separately. The AMI training set consists of 80 hours of data from 137 meetings with 3--5 speakers per meeting recorded by independent headset microphones, and the corresponding reference transcriptions with about 0.9 million words are used for AMI LM training. The merged transcriptions from 
both SWB and Fisher training are combined as the SWB LM training set, which consists of about 27 million words from 14,107 telephone conversations, and the SWB acoustic model training set has around 300 hours of audio. 
For validation, texts from the AMI development (\textit{Dev}) set were used for AMI LM, and about 10\% of the conversations were randomly selected and held out from SWB LM training. The AMI evaluation (\textit{Eval}) set is used as the unseen test set to evaluate the performance of the AMI LMs. For SWB LM evaluation, the \textit{Eval2000} test set which contains conversations from CallHome (CH) and SWB, as well as the \textit{RT03} telephone conversation test set were used.

\subsection{Model Specifications}
Each AMI LM has 8 Transformer blocks with each block taking 512-dimensional (-d) input vectors and having 8 attention heads. The segment length $T$ for the AMI LM is 64 for the standard TLM whereas the segment length is 32 when segment-wise recurrence is used, to ensure that each attention mechanism covers the same sequence length. Each SWB LM uses 24 Transformer blocks with each block taking 512-d input vectors and using 8 attention heads. The segment length for the SWB LMs is 128 for the standard TLM and is 64 when using segment-wise recurrence. The LSTM module of the R-TLM uses a 512-d single-layer unidirectional LSTM in both cases, and the fusion layer outputs 512-d vectors\footnote{\url{https://github.com/BriansIDP/RTLM}}. The R-TLMs with a fusion layer are referred to as f-R-TLMs while those without are referred to as d-R-TLMs, where f- and d- stand for ``fusion'' and ``direct'' respectively. For completeness, LSTM-LMs were also trained and tested for both tasks. For the AMI task, a single-layer LSTM with 512-d hidden states was used, while for the SWB task, a 2-layer LSTM-LM with 2048-d hidden states was used. Similar to the extended history, the LSTM-LM uses the hidden states derived from the 1-best ASR outputs of the history utterances to initialise the LSTM-LM at the beginning of the rescoring stage for each utterance. 

For speech recognition experiments on AMI, a hybrid ASR system with the factorised time delay neural network (TDNN-F) \cite{Ctdnnf} acoustic model was built with the Kaldi toolkit \cite{Ckaldi}, and used lattice-free maximum mutual training \cite{Clfmmi} without any form of data augmentation, speaker adaptation or voice tract length normalisation \cite{AMAMI}. For SWB experiments, the TDNN-F model uses
i-vector speaker adaptation and speed perturbation for data augmentation following the Kaldi recipe. The rescoring with neural network LMs used 100-best lists, where the 100-best lists were generated for each test set using the corresponding 4-gram LMs respectively. LMs were evaluated in terms of both perplexity (PPL) and WER with ASR systems. The statistical significance of WER improvements was evaluated by
the matched pairs sentence segment word error test (MPSSWE) using the NIST ASR scoring toolkit (SCTK).
\section{Experimental Results}
\label{sec:results}
\subsection{AMI Experiments}
The AMI data was first used to search for the most suitable blocks to include the proposed LSTM module. 
It was found that adding the LSTM module to any single transformer block led to improvements but adding to more blocks did not improve the system further. For simplicity, the Transformer block with the best PPL was chosen. As the LSTM module was only added to one Transformer block, increase in training and rescoring time was negligible. The PPL experiments on the AMI Dev set showed that adding the LSTM module in the third block for the fused R-TLM, and in the first block for the direct R-TLM yielded the best results. The PPL and WER results using different LMs on the AMI Eval set are listed in Table \ref{tab:ami_wer}.

\begin{table}[t]
    \centering
    \begin{tabular}{lcc}
    \toprule
       System  &  PPL & WER (\%) \\
       \midrule
       4-gram LM & 88.4 & 20.2 \\
       LSTM-LM cross-utt. & \textbf{56.9} & 18.0 \\
       TLM single-utt. & 74.4 & 18.8 \\
       TLM & 66.8 & 18.4 \\
       \midrule
       TLM XL & 62.8 & 18.3 \\
       R-TLM & 62.3 & 18.1 \\
       d-R-TLM XL & 57.1 & \textbf{17.9} \\
       f-R-TLM XL & 57.3 & \textbf{17.9}\\
       \midrule
       TLM XL + LSTM-LM & 53.8 & 17.7\\
       d-R-TLM XL + LSTM-LM & \textbf{51.4} &  \textbf{17.5} \\
       \bottomrule
    \end{tabular}
    \caption{PPL and WER on AMI Eval set using different LMs. TLM denoted with \textit{single-utt.} refers to rescoring each utterance individually without the extended history. 
    The last two rows present the results for TLM and R-TLM XL interpolated with an LSTM-LM.}
    \label{tab:ami_wer}
    \vspace{-0.5cm}
\end{table}

Although the cross-utterance information in TLM brought 0.4\% absolute WER reduction, and the segment-wise recurrence during training brought a further 0.1\% absolute WER reduction, on small scale text training corpora such as AMI, both PPL and WER for the cross-utterance TLM were worse than those for the LSTM-LM with cross-utterance information. However, when the LSTM module is added to the most suitable Transformer block, both the WER and PPL were reduced to a similar level as the LSTM-LM. By using segment-wise recurrence together with the R-TLM, both direct and fused systems achieved a 0.9\% absolute WER reduction compared to the single-utterance TLM, and a 0.5\% absolute WER reduction compared to the TLM with extended history. 

Moreover, to demonstrate that the R-TLM is distinct from an LSTM-LM, interpolation between the R-TLM and the LSTM-LM was performed. The interpolation weight was fixed at 0.6 throughout the paper. A further WER reduction of 0.4\% absolute compared to the R-TLM XL, and a WER reduction of 0.2\% absolute compared to the TLM XL interpolated with the LSTM-LM, were obtained.

\subsection{Eval2000 and RT03 Experiments}
Experiments were performed on the SWB tasks using both the Eval2000 and RT03 sets for testing. For simplicity, the LSTM module was added to the first Transformer block in all R-TLMs trained on the combined SWB and Fisher corpora. The increase in training and rescoring time was also negligible for the larger model. The PPL and WER results on the Eval2000 set are shown in Table \ref{tab:swbd_wer}.

\begin{table}[t]
    \centering
    \begin{tabular}{lcc}
    \toprule
       System  & PPL & WER (\%) \\
       \midrule
       4-gram LM & 87.2 & 13.1 (9.0 / 17.3)\\
       LSTM-LM cross-utt. & 39.9 & 10.5 (7.0 / 14.0) \\
       TLM single-utt.  & 50.1 & 10.8 (7.2 / 14.4)\\
       TLM & 36.4 & 10.5 (7.0 / 14.0) \\
       \midrule
       TLM XL & \textbf{34.1} & 10.4 (7.0 / 13.8) \\
       R-TLM & 38.5 & 10.5 (6.9 / 13.9) \\
       d-R-TLM XL  & 34.5	& {10.3 (6.9 / 13.6)} \\
       f-R-TLM XL & 34.5	& \textbf{10.2 (6.9 / 13.5)}\\
       \midrule
       TLM XL + LSTM-LM & \textbf{31.9} & 10.1 (6.8 / 13.3) \\
       d-R-TLM XL + LSTM-LM & {32.6} & \textbf{10.0 (6.7 / 13.2)}  \\
       \bottomrule
    \end{tabular}
    \vspace{-0.1cm}
    \caption{The PPLs and WERs on Eval2000.
    The WERs are also split into the SWB and CH parts in the form of (SWB / CH).}
    \label{tab:swbd_wer}
\end{table}
For moderate sized LM training sets, such as the combined SWB and Fisher corpus, the LSTM-LM and TLM with the extended history gave similar performance in WER, while the TLM has a lower PPL. Although the TLM trained with segment-wise recurrence achieves the best PPL in Table~\ref{tab:swbd_wer}, the best WER for single systems is obtained using the R-TLM XL with a fusion layer, which gave a 0.6\% absolute WER reduction compared to the single-utterance TLM, and a 0.2\% absolute WER reduction compared to the TLM with extended history. The lowest WER in the table was achieved by the R-TLM model interpolated with the LSTM-LM. 


It is believed that the R-TLM model has a more robust history representation which is better at modelling less frequent words where acoustic errors are more likely to happen, which explains the fact that TLM XL ended up having the lowest PPL but not the lowest WER. To support this conjecture, negative log-probabilities (\textit{i.e.} LM scores) from different TLMs were measured on error-prone words that appeared in the test set as shown in Table \ref{tab:swbd_ppl}. Error-prone words are here defined as words where more than half of their occurrences in the test set are incorrectly recognised. Additionally, the average LM score was measured for each distinct word appeared in the test set. For words that have lower average LM scores than the TLM with extended history, their occurrences in the training set were averaged by the number of distinct words with lower LM scores to indicate how frequent those words are in the training set. Note that the average occurrence for the TLM is the average training set occurrence of all distinct words in the test set.

\begin{table}[t]
    \centering
    \begin{tabular}{lcc}
    \toprule
       System & Error-prone words & Average occurrences \\
       \midrule
       TLM  & 10.56	& 7,041 \\
       TLM XL & 10.36 & 8,510\\
       R-TLM & 10.31 & 3,428\\
       d-R-TLM XL & \textbf{10.25} & 6,900\\
       \bottomrule
    \end{tabular}
    \vspace{-0.1cm}
    \caption{Average LM scores on error-prone words and average occurrences of words with lower LM scores than the TLM with extended history. The error words column contains average LM scores for incorrectly recognised words in TLM, while the average occurrence indicates how frequent the words with lower LM scores compared to the TLM with extended history are in the training set.}
    \label{tab:swbd_ppl}
    \vspace{-0.5cm}
\end{table}
The TLM XL produces higher LM scores on error words than the R-TLM, indicating those words in the transcriptions are less likely. This explains why the TLM XL produces results with a lower PPL but higher WER than the R-TLM.  
Moreover, the TLM XL is prone to reduce the PPLs of the common words among which recognition errors are less likely to happen. In contrast, adding an LSTM module improves the modelling of less frequent words while maintaining a similar test set PPL as the TLM XL. Therefore, the combined cross-utterance modelling provides more robust representations for ASR systems. 

\begin{table}[t]
    \centering
    \begin{tabular}{lcc}
    \toprule
       System  & Significance & $p$-value \\
       \midrule
       TLM XL & NS & 0.384\\
       R-TLM & NS & 0.472\\
       d-R-TLM XL & S & \textbf{0.010} \\
       f-R-TLM XL  & S & 0.032\\
       \bottomrule
    \end{tabular}
     \vspace{-0.1cm}
    \caption{Significance tests performed on the Eval2000 set where S denotes significance and NS denotes the opposite. Systems  compared to TLM with extended history, and decisions were made at a $p$-value of 0.05.}
    \label{tab:swbd_sig}
\end{table}
Furthermore, to determine if the obtained improvements are statistically significant, MPSSWE was performed as shown in Table \ref{tab:swbd_sig}.
From Table \ref{tab:swbd_sig}, using either segment-wise recurrence or an LSTM module on its own does not generate any significant improvements, whereas the proposed R-TLM with segment-wise recurrence, either using the fusion layer or not, provides statistically significant improvements at a $p$-value of 0.05. 

Finally, LMs were tested on the RT03 evaluation set, and similar improvements to those on Eval2000 set were found as shown in Table \ref{tab:rt03}. The improvements found by using the R-TLM with segment-wise recurrence is significant even at a $p$-value level of 0.001.

\begin{table}[t]
    \centering
    \begin{tabular}{lccc}
    \toprule
       System  & WER (\%) & Significance\\
       \midrule
       4-gram LM & 15.4 (19.2 / 11.5) & -\\
       LSTM-LM & 12.4 (15.7 / 8.8) & - \\
       TLM single-utt.  & 12.7 (16.0 / 9.1) & - \\
       TLM & 12.1 (15.3 / 8.7) & - \\
       \midrule
       TLM XL & 12.0 (15.2 / 8.5) & NS (0.084)\\
       R-TLM & 12.0 (15.2 / 8.6) & NS (0.509)\\
       d-R-TLM XL & \textbf{11.9 (15.0 / 8.5)} & \textbf{S (0.001)}\\
       f-R-TLM XL & \textbf{11.9 (15.0 / 8.5)} & \textbf{S (0.001)}\\
       \midrule
       TLM XL + LSTM-LM & 11.8 (15.0 / 8.3) & - \\
       d-R-TLM XL + LSTM-LM & \textbf{11.8 (14.9 / 8.3)} & - \\
       \bottomrule
    \end{tabular}
    \vspace{-0.2cm}
    \caption{WER on RT03 evaluation set where the same naming convention is used. WER is split into SWB part and Fisher part in the form of (SWB/Fisher). Systems compared to TLM with extended history and  significance decisions  made at $p$-value of 0.05.}
    \vspace{-0.5cm}
    \label{tab:rt03}
\end{table}

\section{Conclusions}
\label{sec:conc}
In this paper, the R-TLM structure is proposed which incorporates an LSTM module into a subset of transformer blocks to obtain more robust and powerful cross-utterance representations. The LSTM module is able to provide a complementary history representation in addition to the segment-wise recurrence. A fusion layer is proposed to connect the LSTM module and the Transformer block. Experiments on the AMI meeting and the SWB conversational transcription tasks showed consistent improvements in WER over three different test sets. Improvements found in SWB task were further supported by significance tests. Moreover, the proposed R-TLM can be interpolated with an LSTM-LM to obtain further reductions in WER.

\bibliographystyle{IEEEbib}
\bibliography{strings,refs}

\end{document}